\documentclass[]{fairmeta}

\usepackage{makecell}
\usepackage{wrapfig}
\usepackage{tabularx}
\usepackage{textcomp}
\usepackage{stfloats}
\usepackage{url}
\usepackage{verbatim}
\usepackage{titlesec}
\usepackage{tocloft}
\usepackage{adjustbox}
\usepackage{multirow}
\usepackage{pifont}
\usepackage[sc]{mathpazo}
\usepackage{tikz}
\usepackage{comment}
\usepackage{amsmath,amssymb}
\usepackage{colortbl}
\usepackage{natbib}
\usepackage{color}
\usepackage{booktabs} 
\usepackage{hyperref}
\usepackage{graphicx}
\usepackage{subcaption}
\usepackage{multirow}
\usepackage{subcaption}

\RequirePackage{xspace}
\makeatletter
\DeclareRobustCommand\onedot{\futurelet\@let@token\@onedot}
\def\@onedot{\ifx\@let@token.\else.\null\fi\xspace}

\def\eg{\emph{e.g}\onedot}

\makeatother

\newcommand{\zhixiangc}[1]{{\color{orange}[Zhixiang: #1]}}

\renewcommand{\paragraph}[1]{\vspace{1.25mm}\noindent\textbf{#1}}

\definecolor{codeblue}{rgb}{0.25, 0.5, 0.5}
\definecolor{codekw}{rgb}{0.35, 0.35, 0.75}
\lstdefinestyle{Pytorch}{
    language = Python,
    backgroundcolor = \color{white},
    basicstyle = \fontsize{9pt}{8pt}\selectfont\ttfamily\bfseries,
    columns = fullflexible,
    aboveskip=1pt,
    belowskip=1pt,
    breaklines = true,
    captionpos = b,
    commentstyle = \color{codeblue},
    keywordstyle = \color{codekw},
}

\definecolor{green}{HTML}{009000}
\definecolor{red}{HTML}{ea4335}

\title{Generative World Renderer}

\author[1,2]{Zheng-Hui Huang}
\author[1,*]{Zhixiang Wang}
\author[1]{Jiaming Tan}
\author[1,3]{Ruihan Yu}
\author[1,3]{Yidan Zhang}
\author[1]{Bo Zheng}
\author[4]{Yu-Lun Liu}
\author[2]{Yung-Yu Chuang}
\author[1]{Kaipeng Zhang}

\affiliation[1]{Alaya Studio, Shanda AI Research Tokyo\\}
\affiliation[2]{National Taiwan University}
\affiliation[3]{The University of Tokyo}
\affiliation[4]{National Yang Ming Chiao Tung University}

\abstract{
Scaling generative inverse and forward rendering to real-world scenarios is bottlenecked by the limited realism and temporal coherence of existing synthetic datasets. To bridge this persistent domain gap, we introduce a large-scale, dynamic dataset curated from visually complex AAA games. Using a novel dual-screen stitched capture method, we extracted \textbf{4M} continuous frames (\textbf{720p/30 FPS}) of synchronized RGB and \textbf{five} G-buffer channels across \textbf{diverse} scenes, visual effects, and environments, including adverse weather and motion-blur variants. This dataset uniquely advances bidirectional rendering: enabling robust in-the-wild geometry and material decomposition, and facilitating high-fidelity G-buffer-guided video generation. Furthermore, to evaluate the real-world performance of inverse rendering without ground truth, we propose a novel VLM-based assessment protocol measuring semantic, spatial, and temporal consistency. Experiments demonstrate that inverse renderers fine-tuned on our data achieve superior cross-dataset generalization and controllable generation, while our VLM evaluation strongly correlates with human judgment. Combined with our toolkit, our forward renderer enables users to edit styles of AAA games from G-buffers using text prompts.
}

\project{\url{https://alaya-studio.github.io/renderer/}}
\code{\url{https://github.com/ShandaAI/AlayaRenderer}}
\correspondence{Zhixiang Wang <\email{zhixiang.wang@shanda.com}>}

\date{\today}

\begin{document}

\maketitle

\begin{figure}[!h]
    \centering
  \includegraphics[width=\linewidth]{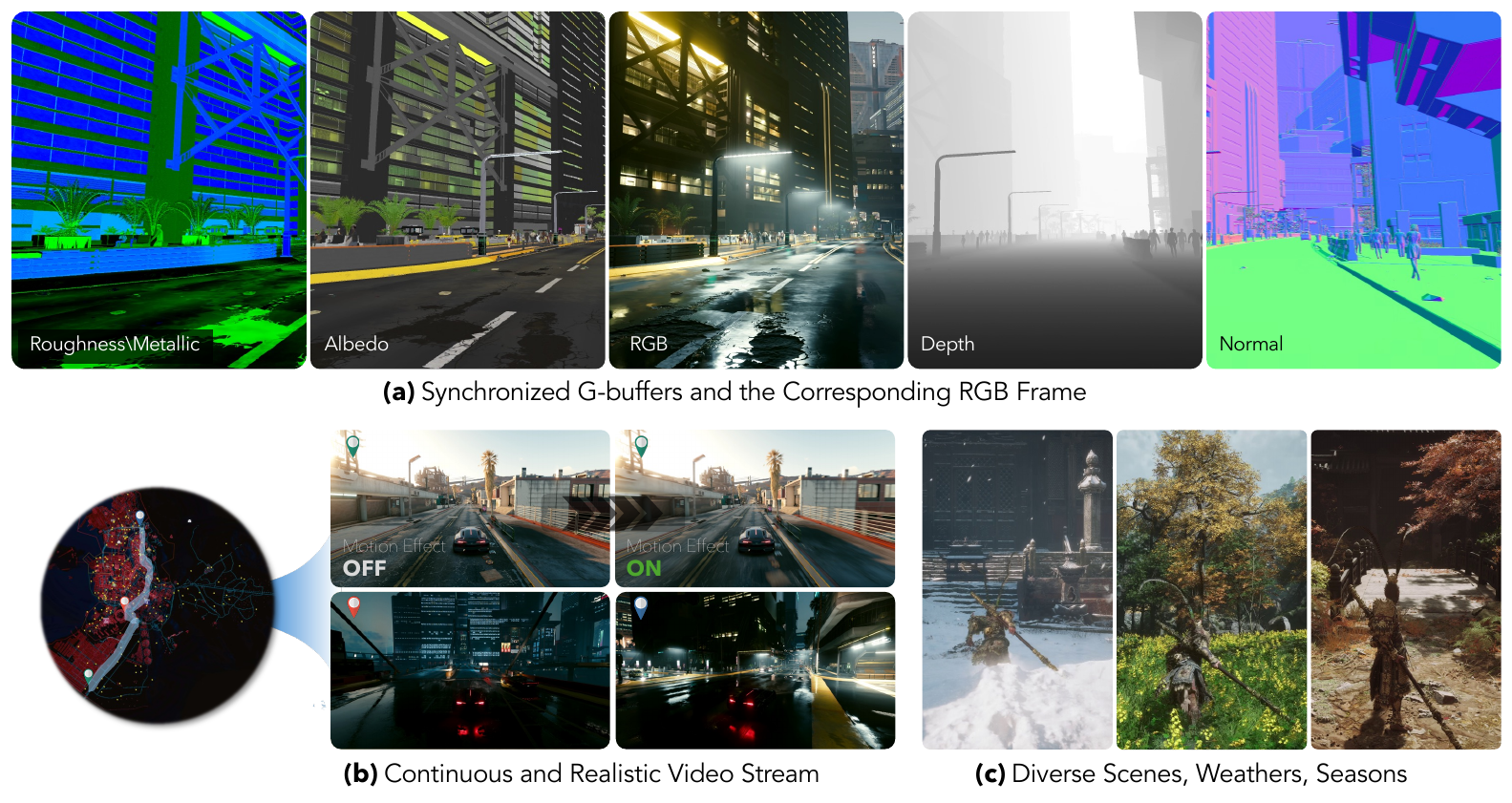}
  \vspace{-5pt}
  \caption{\textbf{We present a large-scale dataset curated from game engines to support scalable generative world rendering.} The dataset provides high-resolution RGB videos with aligned G-buffers, covering continuous and dynamic scenes, long temporal trajectories, and diverse visual conditions. 
  }
  \label{fig:teaser}
\end{figure}

\section{Introduction}

Digital world modeling involves two fundamental tasks: forward rendering, which synthesizes photorealistic images from scene attributes (geometry, materials, and lighting) via the rendering equation~\cite{kajiya1986rendering}; and inverse rendering, which decomposes observed images back into these physical components. 
Recent advancements in generative models have begun to bridge this gap, treating rendering and its inverse as two sides of the same coin within a unified framework~\cite{liang2025diffusion, chen2024unirendere}. At the heart of this unification lies the G-buffer—a rich, intermediate representation that provides explicit geometric and material guidance for controllable synthesis while serving as the supervision target for decomposition.

\begin{figure*}[t] 
    \centering
    \includegraphics[width=\textwidth]{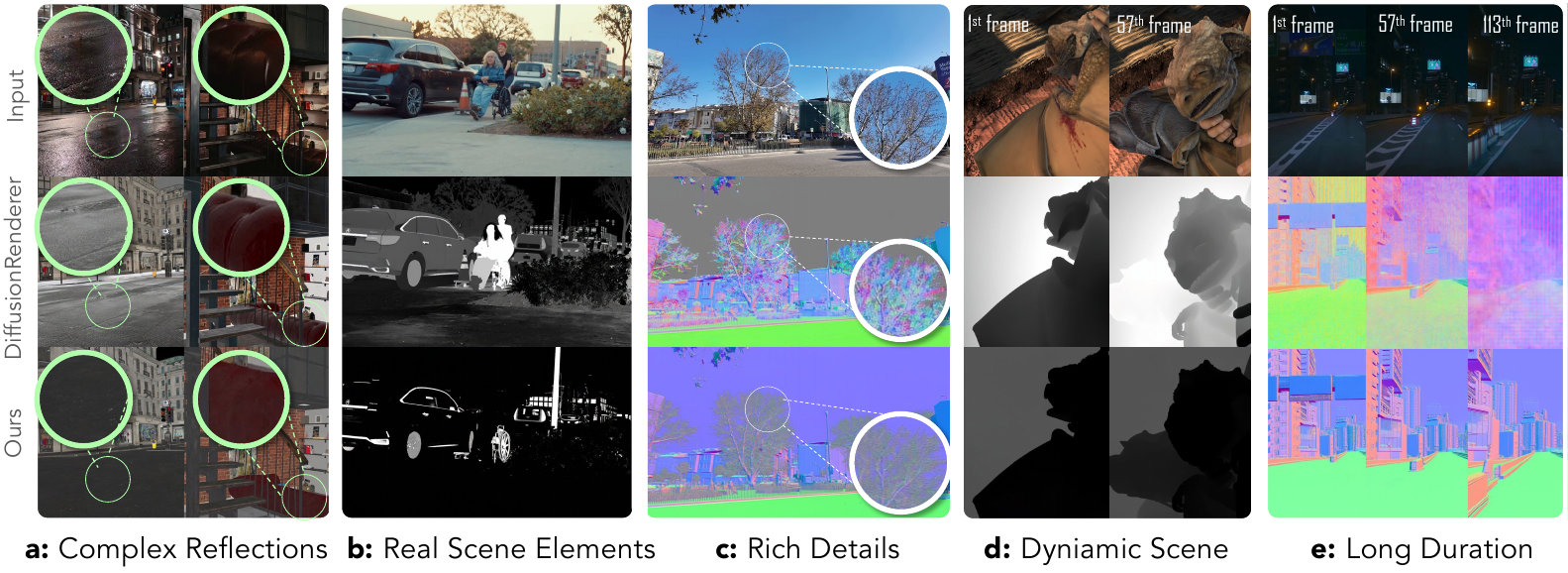}
    \caption{\textbf{Motivation.} Existing approaches, such as DiffusionRenderer, are primarily trained on synthetic datasets, struggling to capture (a) complex reflection and illumination effects, (b) real scene elements (e.g., humans and cars), (c) fine-grained visual details, (d) dynamic motion, and (e) long-range temporal dependencies in long video sequences.
    By contrast, the proposed dataset provides high-fidelity, scene-level supervision with rich geometry, appearance, and temporal dynamics, enabling scalable generative video inverse rendering that generalizes more effectively to real-world scenarios. 
    }
    \label{fig:motivation}
\end{figure*}

Despite the conceptual elegance of this unification, scaling bidirectional rendering to ``in-the-wild'' scenarios remains a formidable challenge. The primary bottleneck is data: the scarcity of large-scale, diverse, and temporally continuous video sequences synchronized with high-fidelity ground-truth G-buffers. 
Existing synthetic datasets often feature limited scene complexity, static camera trajectories, simplified material models, and a lack of adverse weather conditions like fog, rain, or snow. These limitations lead to a persistent domain gap, where models fail to handle the long-tail complexity of real-world videos—such as imperfect "delighting" in cluttered environments, fine-grained vegetation geometry, or temporal flickering under rapid motion.
As illustrated in Figure~\ref{fig:motivation}, these data-starved models struggle to maintain physical plausibility and temporal coherence. This data bottleneck cap the potential of current models to tackle real-world complexity and serve as \textbf{generative world renderers}.

In this paper, we address this data bottleneck by introducing a large-scale, continuous video dataset curated from two AAA games, specifically designed to advance both video inverse rendering and G-buffer-conditioned forward synthesis. 
Our dataset comprises over 4M frames at 720p/30fps, featuring five synchronized G-buffer channels (depth, normals, albedo, metallic, and roughness) aligned with high-quality RGB frames. Unlike previous short-clip collections, our data consists of long, uninterrupted sequences across diverse urban and natural environments under varying atmospheric conditions (e.g., sunny, rainy, foggy, sunset). We develop a non-intrusive pipeline that intercepts runtime G-buffers at the rendering API level, bypassing the need for decompilation or asset extraction. We employ a dual-screen stitched capture strategy to record high-resolution buffers with minimal quality loss.

Crucially, this dataset enables a leap in bidirectional capability. For inverse rendering, it provides the dense supervision necessary for robust material decomposition in complex scenes. For forward rendering, it allows generative models to learn a flexible prior that transcends rigid geometry; for instance, our model can leverage G-buffers to synthesize complex volumetric effects (e.g., fog and rain) that are often omitted in simplified physics-based renderers.

To facilitate wider practical applicability, our dataset pairs clean RGB frames with synthesized motion-blur variants, ensuring that models trained on our data remain resilient to common real-world imaging degradations. Furthermore, recognizing the inherent challenges in evaluating real-world performance, we introduce a VLM-based evaluation protocol. This framework systematically assesses semantic correctness, spatial fidelity, and temporal consistency, demonstrating a strong correlation with human preferences in scenarios where traditional per-frame metrics fall short.

\paragraph{Contributions.} Our contributions are three-fold:
\begin{itemize}
    \item \textbf{A Large-Scale Dataset:} A continuous, high-fidelity G-buffer and video dataset featuring 4M frames with rich dynamics, diverse weather, and long-term temporal coherence.
    \item \textbf{An Efficient Data Curation Pipeline:} A novel capture framework based on graphics API interception and dual-screen stitching that enables scalable acquisition of high-resolution G-buffers.
    \item \textbf{Enhanced Rendering Performance \& Evaluation:} Evidence that fine-tuning on our data significantly improves state-of-the-art models (e.g., DiffusionRenderer) in both decomposition and controllable editing, supported by a new VLM-based ranking protocol for real-world assessment.
\end{itemize}

\section{Related Work}
\label{sec:related}

\paragraph{Inverse Rendering and Forward Rendering Methods.}

Forward rendering synthesizes images from scene attributes by solving
the rendering equation~\cite{kajiya1986rendering}, typically via Monte
Carlo path tracing with microfacet BRDF
models~\cite{cook1982reflectance,walter2007microfacet} as systematized
in modern rendering
pipelines~\cite{pharr2016pbr,akenine2018realtime}.
On the sampling front, spatiotemporal reservoir resampling
(ReSTIR)~\cite{bitterli2020restir,lin2022gris} has dramatically
accelerated real-time path tracing by reusing samples across pixels and
frames, enabling interactive rendering of scenes with millions of
dynamic lights---including the \textit{Cyberpunk 2077} environments from which
our dataset is collected.
The broader neural rendering
paradigm~\cite{tewari2022advances} has progressively augmented or
replaced components of the classical pipeline.
Early works learn to interpret neural textures via deferred
shading~\cite{nalbach2017deep,thies2019deferred} and extend this to
free-viewpoint relighting~\cite{gao2020deferred} and city-scale
lighting factorization~\cite{liu2020factorize};
neural appearance models then compress complex layered SVBRDFs into
compact latent textures decoded by small MLPs for real-time BRDF
evaluation, importance sampling, and level-of-detail
filtering~\cite{rainer2019neural,kuznetsov2021neumip,sztrajman2021neural,fan2022neural,zeltner2024neural,xu2025neuralmaterials};
neural radiance caching accelerates global illumination at path
vertices~\cite{muller2021nrc};
and end-to-end neural renderers such as
RenderFormer~\cite{zeng2025renderformer} directly render triangle
meshes with full global illumination via a transformer, without
per-scene training.
Most recently, diffusion models have been repurposed as data-driven
generative renderers that directly map G-buffers and lighting
descriptions to photorealistic
images~\cite{liang2025diffusion,chen2024unirendere,zeng2024rgb},
or augment the forward pass with physics-inspired
constraints~\cite{pilight2026}.
Unlike classical path tracers, these generative approaches implicitly
learn complex light transport---including volumetric scattering,
global illumination, and view-dependent effects---from paired
supervision, bypassing explicit material models and costly per-sample
Monte Carlo integration.
This paradigm also extends to controllable relighting and scene
editing~\cite{jin2024neural,zhang2025scaling,he2025unirelight,physicalrelighting2025},
yet its scalability is fundamentally gated by the availability of
large-scale, temporally continuous G-buffer--RGB pairs---precisely the
gap our dataset addresses.

Inverse rendering decomposes images into geometry, reflectance, and materials for relighting and editing. Early optimization methods~\cite{land1971lightness,barron2014shape,grosse2009ground} struggle with real-world complexity. Learning-based approaches improved intrinsic decomposition~\cite{li2018cgintrinsics,sengupta2019neural,li2020inverse,li2021openrooms,careaga2024colorful} and material estimation~\cite{li2018materials,deschaintre2018single,boss2020two,lopes2024material} using synthetic supervision, while neural fields enabled joint optimization via shape-reflectance factorization~\cite{zhang2021nerfactor}, tensorial representations~\cite{jin2023tensoir}, room-scale decomposition~\cite{ye2023intrinsicnerf}, neural SDFs~\cite{zhu2023i2}, Gaussian splatting with BRDF decomposition~\cite{liang2024gs,gao2024relightable,saito2024relightable}, and physics-based losses~\cite{wu2025pbr}---though all require per-scene optimization. Diffusion models improved generalization through joint intrinsic prediction~\cite{luo2024intrinsicdiffusion}, bidirectional material decomposition~\cite{zeng2024rgb}, probabilistic formulations~\cite{kocsis2024intrinsic}, stochastic inverse rendering~\cite{enyo2024diffusion}, lighting-material disambiguation~\cite{chen2024intrinsicanything}, multi-view intrinsic decomposition~\cite{li2024idarb}, SVBRDF synthesis~\cite{vecchio2024matfuse,sartor2023matfusion,vecchio2024controlmat}, single-image PBR extraction~\cite{lopes2024material,litman2025materialfusion}, text-to-PBR generation~\cite{kocsis2025intrinsix}, and multi-view G-buffer estimation~\cite{wang2025mage,he2025neural}. On the relighting side, diffusion-based methods enable object-level relighting~\cite{jin2024neural}, illumination editing with consistent light transport~\cite{zhang2025scaling}, indoor scene relighting~\cite{xing2025luminet}, video relighting~\cite{he2025unirelight}, and dynamic human rendering~\cite{wang2024intrinsicavatar}. Recent video diffusion models enable temporally consistent inverse rendering~\cite{liang2025diffusion} and video-level PBR material extraction~\cite{munkberg2025videomat}, but train on \emph{short, object-centric} clips---our dataset provides the \emph{long, continuous} sequences with complex dynamics these methods require.

\paragraph{Datasets and Game-Based Collection.}
Acquiring real-world G-buffers remains challenging, making synthetic datasets essential. Indoor datasets provide disentangled reflectance~\cite{roberts2021hypersim}, SVBRDF annotations~\cite{li2021openrooms,zhu2022learning}, laser scans~\cite{yeshwanth2023scannet++}, scalable mid-level vision data~\cite{eftekhar2021omnidata}, and controllable lighting at scale~\cite{li2018interiornet}. Material resources offer PBR collections~\cite{vecchio2024matsynth,ma2023opensvbrdf,zhou2023photomat}, relighting benchmarks~\cite{ummenhofer2024objects,ren2022diligent102}, while outdoor datasets cover city-scale scenes~\cite{li2023matrixcity,wang2025lightcity}, aerial imagery~\cite{liu2021urbanscene3d}, and driving scenarios~\cite{barua2025gta}. Object-centric resources include video frames~\cite{ling2024dl3dv}, 3D objects~\cite{deitke2023objaverse,wu2023omniobject3d}, outdoor 3DGS data~\cite{xiong2024gauu}, and depth/flow benchmarks~\cite{wang2020tartanair,mehl2023spring}. Procedurally generated datasets~\cite{raistrick2024infinigen,raistrick2024infinigen} provide multi-modal ground truth (depth, normals, albedo) without external assets, and configurable simulation platforms~\cite{ge2024behavior} offer adjustable scene parameters, though both lack the visual fidelity and content diversity of artist-crafted game worlds. Long synthetic video datasets with multi-modal annotations~\cite{zheng2023pointodyssey,yang2024depth} further demonstrate the value of temporal supervision at scale. Games enable photorealistic data extraction via graphics interception~\cite{richter2016playing,richter2017playing}, DirectX injection~\cite{krahenbuhl2018free}, engine plugins~\cite{qiu2017unrealcv,ros2016synthia,huang2018deepmvs,dosovitskiy2017carla,shah2017airsim,pollok2019unrealgt}, and ReShade/OBS pipelines~\cite{zhou2025omniworld}, with domain adaptation addressing sim-to-real gaps~\cite{tobin2017domain,tremblay2018training,hoyer2022daformer,mikami2021scaling}. However, existing datasets remain \emph{image-centric} or provide \emph{sparse channels} with short sequences---we extract \emph{synchronized multi-channel G-buffers} (depth, normals, albedo, metallic, roughness) as continuous long-duration video from AAA games.

\paragraph{Temporal Consistency and Depth Estimation.}
Real videos exhibit motion blur from finite exposure; MPI-Sintel's clean or degraded passes~\cite{butler2012naturalistic} and high-resolution synthetic benchmarks~\cite{mehl2023spring} establish the design philosophy we follow, generating blur via frame interpolation~\cite{huang2022real,jiang2018super,reda2022film}. Temporal consistency methods span recurrent networks~\cite{lai2018learning}, deep video priors~\cite{lei2020blind}, video diffusion~\cite{blattmann2023align,blattmann2023stable}, feature propagation~\cite{geyer2023tokenflow,qi2023fatezero}, spatial-temporal constraints~\cite{yang2024fresco,yang2023rerender}, token merging~\cite{li2024vidtome}, temporal transformers~\cite{yan2023temporally}, content deformation fields~\cite{ouyang2024codef}, flow-based methods~\cite{liang2024flowvid}, streaming video translation~\cite{liang2024looking}, and correspondence-guided diffusion~\cite{chu2024medm}. Depth estimation has advanced with foundation models~\cite{yang2024depth}, diffusion priors~\cite{ke2024repurposing,bochkovskii2024depth,he2024lotus}, joint depth-normal prediction~\cite{hu2024metric3d,fu2024geowizard}, surface normal estimation~\cite{bae2024rethinking,ye2024stablenormal}, optical flow~\cite{wang2024sea,dong2024memflow}, and temporally consistent video depth~\cite{chen2025video,hu2025depthcrafter,shao2025learning,wang2024nvds}. Recent work further extends video diffusion priors to temporally consistent normal estimation~\cite{bin2025normalcrafter}. Our long sequences with ground-truth geometry support training and validating such temporally-aware methods.

\paragraph{Evaluation Protocols.}
Standard metrics (PSNR, LPIPS~\cite{zhang2018unreasonable}) miss cross-buffer consistency, while perceptual metrics~\cite{fu2023dreamsim} and video metrics (FVD~\cite{unterthiner2019fvd}) show quality-consistency trade-offs~\cite{ge2024content,huang2024vbench}. For real videos without ground truth, VLMs enable semantic evaluation via quality assessment~\cite{wang2023exploring,wu2023q,wu2023qalign}, faithfulness VQA~\cite{hu2023tifa}, compositional benchmarks~\cite{huang2023t2i}, 3D evaluation~\cite{wu2024gpt}, preference learning~\cite{xu2023imagereward}, and multimodal depiction~\cite{deqa_score,depictqa_v2,depictqa_v1}. More broadly, the LLM-as-a-Judge paradigm~\cite{zheng2023judging,liu2023g} has been extended to vision-language settings~\cite{chen2024mllm,lee2024prometheus}, video quality understanding~\cite{he2024videoscore,zhang2024q,wang2025aigv}, and open evaluation platforms for generative models~\cite{jiang2024genai}. We introduce a VLM-based ranking protocol targeting material channels (metallic, roughness) where VLM priors provide meaningful common-sense judgments.

\begin{figure*}[t] 
    \centering
    \includegraphics[width=\textwidth]{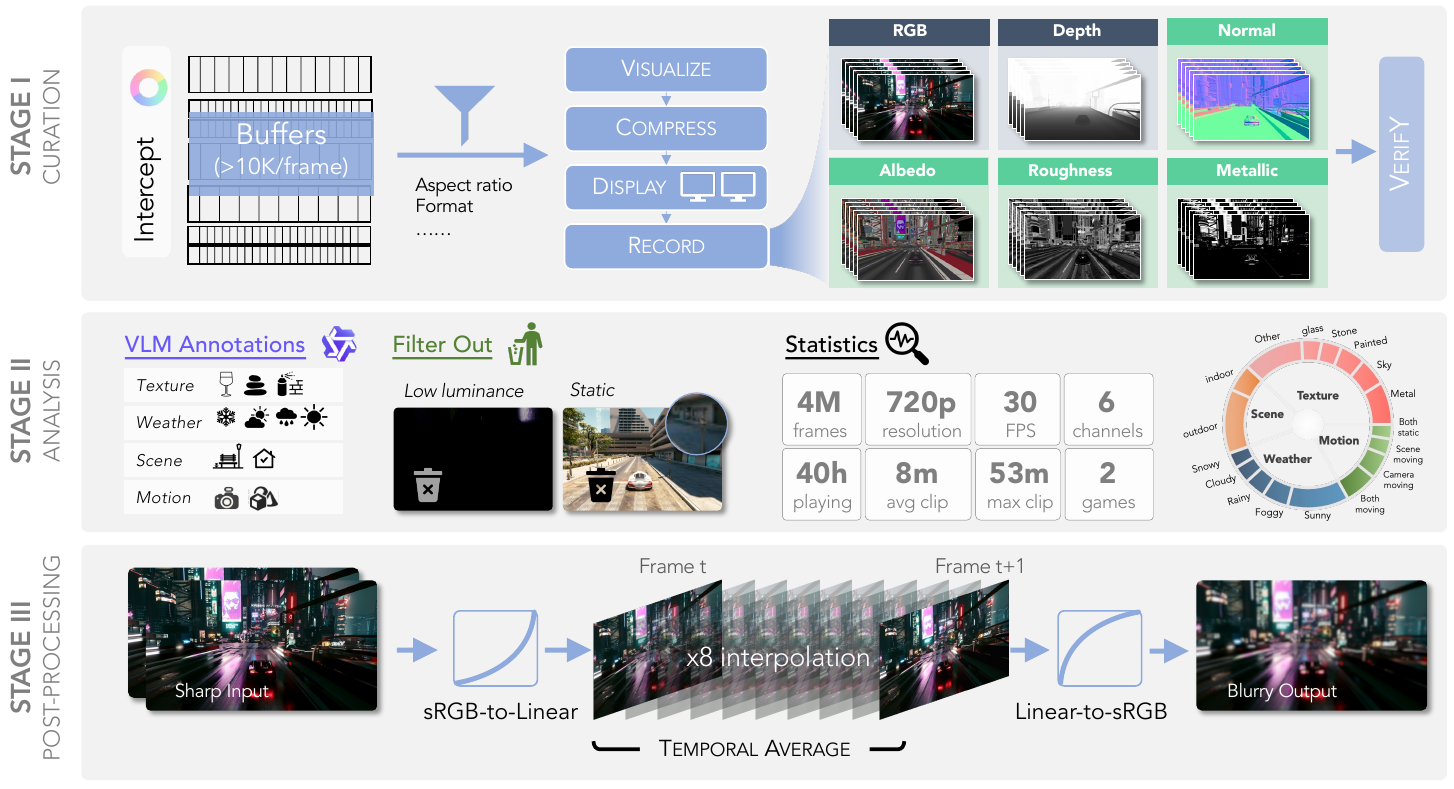}
    \caption{\textbf{Pipeline.} \textbf{Stage I}: We curate video sequences containing RGB frames and five corresponding G-buffer channels from commercial game engines. Buffer interception is performed using ReShade, which allows us to capture intermediate rendering outputs at runtime. Because a single rendering pass exposes thousands of heterogeneous and largely irrelevant buffers, an automated filtering procedure is required to identify valid candidates.
    To disambiguate target G-buffers from irrelevant render targets, we utilize RenderDoc for offline inspection to define filtering rules based on metadata invariants. By manually validating the semantic accuracy of the remaining buffers, we iteratively optimize these rules into robust signatures that ensure consistent runtime identification. As a final verification step, we re-render the RGB frames from the collected G-buffers using a deferred shading pipeline and check for pixel-level consistency with the original RGB outputs. 
    \textbf{Stage II}: We annotate meta-information for each sequence and filter out unsatisfactory frames based on quality criteria.
    \textbf{Stage III}: We synthesize motion blur to enhance temporal realism and better match real-world capture conditions. 
    }
    \vspace{-4mm}
    \label{fig:pipeline}
\end{figure*}

\section{Dataset Construction}

\subsection{G-buffer Interception}
We utilize ReShade~\cite{reshade} to intercept the rendering pipeline at the graphics API level. 
However, extracting complete G-buffers is non-trivial because modern games employ engine- and title-specific G-buffer
packing/encodings, with no standardized layout across titles. 
To address this, we first conduct offline frame analysis with RenderDoc~\cite{renderdoc} to identify the candidate
render passes and their associated render-target attachments, including their formats, dimensions, sample counts.

Based on offline frame inspection and pass tracing, we implement game-specific ReShade add-ons that hook graphics API callbacks to monitor per-frame render-target bindings. During runtime, we maintain a small pool of candidate attachments and GPU-copy only those that satisfy stable invariants, including consistent format as well as extent and recurrent binding patterns. To disambiguate the desired render target, we log lightweight per-frame signatures, including format and extent stability and pass-local draw-call spans. After selection, we bind the selected render target as an input texture to the ReShade effect runtime; the effect shader can then shade it to the screen.

In our capture pipeline, the only normal information we can reliably obtain from the rendering pipeline is a world-space normal buffer. We adopt camera-space normals to match prior inverse rendering work~\cite{liang2025diffusion}, but cannot convert world to camera space due to the lack of reliable access to the view matrix. We therefore reconstruct camera-space normals from depth by inverse projection and finite differences during the ReShade effects stage:
\begin{equation}
\mathbf{n}=\operatorname{normalize}\!\left(\frac{\partial \mathbf{P}}{\partial x}\times\frac{\partial \mathbf{P}}{\partial y}\right),
\end{equation}
where $\mathbf{P}$ is the view-space position reconstructed from the depth buffer.

Beyond the geometry, exporting material channels introduces additional reliability challenges. Metallic and roughness are often packed into different channels of a single G-buffer render target. Recording these channels via screen-capture video can cause channel coupling artifacts (e.g., inter-channel bleeding) even under near-lossless settings. We therefore decouple the maps and render them into spatially distinct screen regions, ensuring that compression noise does not cross-contaminate material properties.

\subsection{Synchronized Multi-Screen Recording}
Directly exporting multi-channel G-buffers per frame is prohibitively costly due to storage bandwidth, file-management overhead, and GPU-to-CPU readback stalls. Instead, we shade target buffers to the screen and record them with hardware-accelerated capture, enabling scalable, temporally synchronized acquisition without modifying the game engine.

To ensure strict temporal synchronization across all six data channels, we employ a ``mosaic'' compositing strategy. We render all G-buffers onto a unified canvas captured via OBS (Open Broadcaster Software) at a near-lossless bitrate. To overcome the resolution limits of a single display, we stitch two 2K monitors, allowing us to record each channel at an effective resolution of 720p. Since expanding the display area inherently increases the game's field of view, we apply a center-crop to the source buffers before tiling them onto the final output, preserving the intended aspect ratio.

\subsection{Scene Traversal Strategies}
We collect data from \textit{Cyberpunk 2077} and \textit{Black Myth: Wukong} with distinct traversal strategies to maximize diversity.
For \textit{Cyberpunk 2077}, we use a semi-automated driving setup and define long-range waypoints to generate continuous trajectories with variable speeds, yielding rich temporal dynamics. Beyond driving, we also capture sequences where the player walks along streets and collects indoor scenes to broaden the coverage of viewpoints and environments.
For \textit{Black Myth: Wukong}, we capture exploration sequences from completed save files. We deliberately avoid combat and instead traverse a wide range of environments and routes to obtain diverse appearances and scene content.



\subsection{Dataset Statistics}
We provide rich annotations for the dataset, covering scene, weather, camera and scene motion, and texture. We first deploy Qwen3-VL-235B-A22B-Instruct~\cite{Qwen3-VL} using vLLM~\cite{kwon2023efficient}. For each video clip, we uniformly sample five frames along the temporal axis and feed the timestamped frames into Qwen to obtain the corresponding annotations.

\paragraph{Annotation labels.} For each clip, we annotate four categorical attributes. The \textit{texture} summarizes the dominant material/appearance cues (e.g., plastic, metal, brick, sky, painted, stone, glass, \dots); the weather labels of clips include {sunny, cloudy, foggy, rainy, snowy}; \textit{scene} indicates whether the clip is indoor or outdoor; and \textit{motion} describes camera--scene dynamics with four cases: {camera static scene moving, camera moving scene moving, camera moving scene static, and both static}. Together, these attributes capture diverse environmental settings and visual characteristics, highlighting the broad variability of our dataset across conditions and content.

\paragraph{Distribution Analysis.} We analyze the distributions of metallic and roughness in the two game sources. As shown in Figure~\ref{fig:pipeline}, \textit{Cyberpunk 2077} exhibits a higher proportion of pixels with large metallic values than \textit{Black Myth: Wukong}, while \textit{Black Myth: Wukong} contains more high-roughness regions. This trend is consistent with their dominant visual themes: \textit{Cyberpunk 2077} features metal-rich urban environments, whereas \textit{Black Myth: Wukong} more often depicts natural scenes with rough, diffuse materials. Together, these two domains provide complementary coverage of common real-world material appearances. We further analyze pixel brightness using luminance and the HSV Value channel. Figure~\ref{fig:pipeline} shows that \textit{Black Myth: Wukong} concentrates at low values, consistent with outdoor scenes where occlusions yield more shadowed regions.
In contrast, \textit{Cyberpunk 2077} is more balanced across the range.

Finally, we filter out clips where both the scene content and the camera remain static throughout the annotation, and we exclude frames with excessively low luminance.

\subsection{Dataset Post-processing}
To maximize dataset reusability for downstream pixel-aligned tasks, we capture RGB frames with the engine motion blur disabled, providing sharp canonical observations with clean temporal correspondences. Since real videos often contain camera-induced motion blur, we additionally release an offline motion-blurred RGB variant to reduce this domain gap.

We approximate exposure integration by first interpolating $8$ RGB sub-frames with RIFE~\cite{huang2022real}, averaging them in the linear domain, and converting back to RGB:
\begin{equation}
I^{\text{blur}}_t
= \mathrm{RGB}\!\Big(\tfrac{1}{K}\sum_{i=1}^{K}\mathrm{Lin}\!\big(\tilde{I}_{t,i}\big)\Big),
\end{equation}
where $\tilde{I}_{t,i}$ denotes the RIFE-interpolated RGB frames, and $\mathrm{Lin}(\cdot)$\,/\,$\mathrm{RGB}(\cdot)$ are the RGB$\leftrightarrow$linear conversions.
\section{VLM-based Evaluation on Real-Scene Test Cases}
\label{sec:vlm_eval}

Validating that our dataset improves real-world generalization requires evaluating material predictions on real captures, where ground truth is generally unavailable. User studies offer one alternative, but they scale poorly: judging material properties in complex scenes demands domain expertise and substantial annotation effort.

In contrast, modern vision-language models (VLMs) encode extensive material-related world knowledge and can serve as scalable judges for relative comparisons without requiring ground truth. We focus on metallic and roughness, as these properties exhibit strong semantic and appearance priors (e.g., recognizable material categories and characteristic specular behavior), which facilitate more consistent pairwise preferences. Moreover, for complex videos, VLMs can leverage global context while still attending to localized cues (e.g., thin structures, specular trims, or brief temporal flicker), enabling them to identify subtle but systematic failure patterns across different methods. We provide the detailed prompt used for VLM evaluation in Figure~\ref{fig:prompt}.



\section{Experiments}
\subsection{Training and Experimental Setup}
Our primary baseline is DiffusionRenderer~\cite{liang2025diffusion}, as it is currently the only accessible method that performs video inverse rendering. We additionally include two recent diffusion-based image inverse rendering models~\cite{zeng2024rgb}~\cite{zheng2025dnf} as baselines. Since DiffusionRenderer does not release its training dataset, we cannot reproduce its original training data; we therefore fine-tune the official implementation from the released pre-trained weights to assess whether our dataset improves real-video generalization while keeping the model fixed. We use the Cosmos-based DiffusionRenderer checkpoint as our baseline, as it performs better than the SVD variant. We fully fine-tune the model using fixed-length clips of 57 frames sampled at 24 FPS and a resolution of $1280 \times 720$. We use Cyberpunk 2077 for training and Black Myth: Wukong for testing. We fine-tune two variants on data with and without motion effects, and select the motion-augmented variant as our final model due to its consistently better performance. We also fine-tune a longer-clip variant with 113 frames under the same settings; as shown in Figure~\ref{fig:motivation}, it substantially improves long-video inference. During inference, we strictly follow DiffusionRenderer's original protocol for consistency. For image-based inverse rendering methods, we run inference on each video frame independently.

We also demonstrate game editing as a practical application of our dataset. To implement this, we first utilize Qwen3-VL-235B-A22B-Instruct~\cite{Qwen3-VL} to generate descriptive captions for each video clip. Given that G-buffers provide dense geometric and material priors, our prompts focus exclusively on lighting and environmental effects, enabling users to manipulate these attributes via text during inference. Architecturally, we adapt the Wan 2.1-T2V-1.3B~\cite{wan2025} by incorporating G-buffers as conditional inputs. Following the original training configuration of the base model, we fully fine-tune it on Black Myth: Wukong at a spatial resolution of $832 \times 480$ (480p) and a frame rate of 16 FPS, utilizing 81-frame clips. Evaluation and GeneralizationIn the absence of directly comparable methods for this specific task, we establish a baseline by adapting DiffusionRenderer’s forward renderer. Specifically, we employ DiffusionLight to extract environment maps from the videos, which then serve as the lighting conditions to produce the rendered results. A subset of the Black Myth: Wukong dataset is reserved for testing. Furthermore, to assess the robust generalizability of our model, we conduct cross-dataset evaluations on Cyberpunk 2077. This experiment demonstrates that our model generalizes effectively to unseen game environments, maintaining high-fidelity and controllable video synthesis.

\begin{table*}[t]
\centering
\small
\setlength{\tabcolsep}{4pt}
\renewcommand{\arraystretch}{1.1}
\caption{\textbf{Quantitative evaluation of inverse rendering on black myth video dataset.} DNF denotes DNF-intrinsic, and DR denotes DiffusionRenderer. Note that RGB$\leftrightarrow$X does not output depth.}
\label{tab:main_metrics}
\resizebox{\textwidth}{!}{
\begin{tabular}{l|cccccc|cc|cccc|cc|cc}
\toprule
 & \multicolumn{6}{c|}{\textbf{Depth}} & \multicolumn{2}{c|}{\textbf{Normals}} & \multicolumn{4}{c|}{\textbf{Albedo}} & \multicolumn{2}{c|}{\textbf{Metallic}} & \multicolumn{2}{c}{\textbf{Roughness}} \\
\multicolumn{1}{c|}{} &
\rotatebox{45}{Abs Rel$\downarrow$} & 
\rotatebox{45}{RMSE$\downarrow$} & 
\rotatebox{45}{RMSE log$\downarrow$} &
\rotatebox{45}{$\delta < 1.25$ $\uparrow$} & 
\rotatebox{45}{$\delta < 1.25^2$ $\uparrow$} & 
\rotatebox{45}{$\delta < 1.25^3$ $\uparrow$} & 
\rotatebox{45}{AngularError$\downarrow$} & 
\rotatebox{45}{Acc@11.25$^\circ$ $\uparrow$} &
\rotatebox{45}{PSNR $\uparrow$} & 
\rotatebox{45}{LPIPS $\downarrow$} & 
\rotatebox{45}{si-PSNR $\uparrow$} & 
\rotatebox{45}{si-LPIPS $\downarrow$} &
\rotatebox{45}{RMSE $\downarrow$} & 
\rotatebox{45}{MAE $\downarrow$} & 
\rotatebox{45}{RMSE $\downarrow$} & 
\rotatebox{45}{MAE $\downarrow$}\\
\midrule
RGB$\leftrightarrow$X & - & - & - & - & - & - & 78.05$^\circ$ & 0.035 & 8.74 & \textbf{0.619} & \underline{20.11} & \textbf{0.626} & 0.510 & 0.503 & 0.349 & 0.313 \\
DNF & \underline{0.862} & \underline{0.026} & 0.918 & \underline{0.361} & \underline{0.610} & \underline{0.762} & 53.21$^\circ$ & 0.065 & 13.84 & 0.702 & 15.59 & 0.701 & 0.245 & 0.183 & 0.566 & 0.543 \\
DR & 1.118 & 0.030 & \underline{0.723} & 0.267 & 0.496 & 0.684 & \underline{45.01$^\circ$} & \underline{0.110} & \textbf{17.53} & 0.648 & 19.90 & 0.646 & \underline{0.230} & \underline{0.134} & \underline{0.281} & \underline{0.237} \\
Ours & \textbf{0.697} & \textbf{0.023} & \textbf{0.430} & \textbf{0.609} & \textbf{0.761} & \textbf{0.852} & \textbf{42.57$^\circ$} & \textbf{0.150} & \underline{16.44} & \underline{0.628} & \textbf{21.44} & \underline{0.635} & \textbf{0.104} & \textbf{0.024} & \textbf{0.266} & \textbf{0.218} \\
\bottomrule
\end{tabular}
}
\end{table*}



\begin{table*}[t]
\centering
\small
\setlength{\tabcolsep}{5pt}
\renewcommand{\arraystretch}{1.1}
\caption{\textbf{Quantitative evaluation of inverse rendering on the Sintel dataset.}}
\label{tab:sintel}
\resizebox{0.75\textwidth}{!}
{
\begin{tabular}{l|ccccc|cccc}
\toprule
 & \multicolumn{5}{c|}{\textbf{Depth}} & \multicolumn{4}{c}{\textbf{Albedo}} \\
\multicolumn{1}{c|}{} &
\rotatebox{45}{RMSE$\downarrow$} & \rotatebox{45}{RMSE log$\downarrow$} &
\rotatebox{45}{$\delta < 1.25\uparrow$} & \rotatebox{45}{$\delta < 1.25^2\uparrow$} & \rotatebox{45}{$\delta < 1.25^3\uparrow$} & \rotatebox{45}{PSNR $\uparrow$} & \rotatebox{45}{LPIPS $\downarrow$} & \rotatebox{45}{si-PSNR $\uparrow$} & \rotatebox{45}{si-LPIPS $\downarrow$}\\
\midrule
RGB$\leftrightarrow$X   & - & - & - & - & - & 8.69 & 0.613 & 16.69 & 0.595 \\
DNF-Intrinsic   & \underline{0.249} & 1.090 & \underline{0.371} & \underline{0.590} & \underline{0.710} & 13.16 & 0.546 & 13.68 & 0.535 \\
DiffusionRenderer       & 0.268 & \underline{0.911} & 0.331 & 0.560 & 0.707 & \underline{14.87} & \underline{0.505} & \underline{17.46} & \underline{0.497} \\
Ours   & \textbf{0.220} & \textbf{0.745} & \textbf{0.478} & \textbf{0.649} & \textbf{0.776} & \textbf{15.40} & \textbf{0.486} & \textbf{17.80} & \textbf{0.491} \\
\bottomrule
\end{tabular}
}
\end{table*}




\begin{table}[t]
\begin{minipage}[t]{0.5\linewidth}
\centering
\scriptsize
\setlength{\tabcolsep}{6pt}
\renewcommand{\arraystretch}{1.1}
\caption{\textbf{VLM evaluation metrics.} R: Roughness; M: Metallic.}
\label{tab:ranking_metrics}
\begin{tabular}{cl|ccc}
\toprule
\multicolumn{2}{c|}{Methods} & Sem.$\downarrow$ & App.$\downarrow$ & Temp.$\downarrow$ \\
\midrule
\multirow{3}{*}{\rotatebox[origin=c]{90}{\scriptsize\textbf{R}}}
& DiffusionRenderer        & 2.45 & 2.40 & 2.10 \\
& Ours                     & 1.78 & \textbf{1.78} & 2.08 \\
& Ours (w/ motion blur)             & \textbf{1.78} & 1.83 & \textbf{1.83} \\
\midrule
\multirow{3}{*}{\rotatebox[origin=c]{90}{\scriptsize\textbf{M}}}
& DiffusionRenderer        & 2.35 & 2.28 & 2.00 \\
& Ours                     & 1.90 & 2.13 & 2.15 \\
& Ours (w/ motion blur)             & \textbf{1.75} & \textbf{1.60} & \textbf{1.85} \\
\bottomrule
\end{tabular}
\end{minipage}\hfill
\begin{minipage}[t]{0.48\linewidth}
\centering
\scriptsize
\renewcommand{\arraystretch}{0.9}
\caption{\textbf{User study.} Groups 1 and 2 represent samples where the VLM prefers our model and DiffusionRenderer, respectively. The reported percentages denote the agreement rate between human experts and VLM predictions.}
\label{tab:user_study}
\begin{tabular}{lcc}
\toprule
\textbf{Channel} & \textbf{Group 1} & \textbf{Group 2} \\
\cmidrule(lr){2-2}\cmidrule(lr){3-3}
& prefer our model & prefer DiffusionRenderer\\
\midrule
Metallic  & 85\% & 70\% \\
Roughness & 75\% & 61\% \\
\bottomrule
\end{tabular}
\end{minipage}
\end{table}

\begin{table}[t]
\centering
\scriptsize
\setlength{\tabcolsep}{3pt} 
\renewcommand{\arraystretch}{1.1}
\caption{\textbf{Ablation study on motion blur.}}
\label{tab:ablation_mb}
{ 
\begin{tabular}{l|cccc|cccc}
\toprule
& \multicolumn{4}{c|}{\textbf{Depth}} & \multicolumn{4}{c}{\textbf{Albedo}} \\ 
\cmidrule(lr){2-5} \cmidrule(lr){6-9}
Method & RMSE log$\downarrow$ & $\delta \!<\! 1.25 \uparrow$ & $\delta \!<\! 1.25^2 \uparrow$ & $\delta \!<\! 1.25^3 \uparrow$ & PSNR $\uparrow$ & LPIPS $\downarrow$ & si-PSNR $\uparrow$ & si-LPIPS $\downarrow$ \\
\midrule
Ours & 0.773 & 0.467 & \textbf{0.649} & 0.756 & \textbf{15.73} & 0.513 & 17.37 & 0.513 \\
Ours (w/ motion blur) & \textbf{0.745} & \textbf{0.478} & \textbf{0.649} & \textbf{0.776} & 15.40 & \textbf{0.486} & \textbf{17.80} & \textbf{0.491} \\
\bottomrule
\end{tabular}
}
\end{table}

\subsection{Quantitative Evaluation of Inverse Rendering}

\paragraph{Evaluation Metrics.}
For synthetic benchmarks with ground truth, we report standard metrics for each modality. Following the prior work~\cite{luo2020consistent}, we evaluate depth in disparity space and apply scale-and-shift alignment, reporting AbsRel, RMSE, RMSE-log, and threshold accuracies $\delta < 1.25^n$ ($n\!=\!1,2,3$). For albedo, we report PSNR and LPIPS, as well as their scale-invariant counterparts (si-PSNR and si-LPIPS) to reduce sensitivity to global intensity scaling. For normals, we report mean angular error and the accuracy under an $11.25^\circ$ threshold (Acc@11.25$^\circ$). For material parameters (metallic and roughness), we report RMSE and MAE.


\paragraph{Black Myth Wukong Benchmark.}
As there is currently no public benchmark for video inverse rendering, we construct a quantitative test set from our Black Myth: Wukong capture.
Specifically, we hold out 39 video clips, each containing 57 frames, covering diverse materials, lighting, and dynamic events.
We evaluate all methods on the same held-out clips and report per-modality metrics averaged over frames and then over clips. Our fine-tuned model achieves the best performance on depth and normal estimation, and attains the strongest scale-invariant albedo scores while markedly improving metallic and roughness accuracy in Table~\ref{tab:main_metrics}.

\paragraph{Sintel Benchmark.}
We additionally evaluate on the MPI-Sintel final pass, which contains realistic effects such as motion blur and depth-of-field, and provides ground-truth albedo and depth.
A subtle difference is that Sintel's albedo annotation does not enforce a fully-black sky region, which is inconsistent with our dataset convention.
To avoid penalizing methods for this annotation mismatch, we exclude samples containing sky regions when evaluating albedo on Sintel. As shown in Table~\ref{tab:sintel}, our fine-tuned model achieves the best overall performance on both depth and albedo, improving RMSE/RMSE-log and $\delta$ accuracies for depth while also yielding higher PSNR and lower LPIPS including scale-invariant variants for albedo.

\paragraph{Real-World Video Evaluation.}
To evaluate real-scene generalization, we collect 40 real-world video cases from online sources, spanning indoor and outdoor scenes, varying motion magnitude (slow to fast camera/object motion), and diverse times of day.

Because real videos lack ground-truth intrinsic buffers, we use a video-capable vision-language model (VLM) to score and rank predictions.
We adopt Gemini 3 Pro~\cite{team2023gemini} as the judge model due to its strong video understanding and temporal reasoning.
Given each test video, we compose a fixed-layout grid where the RGB reference and method outputs are synchronously played, and prompt the VLM to rate: (i) temporal consistency, (ii) spatial quality, and (iii) semantic plausibility, producing a structured score and ranking for each modality. We report aggregated VLM rankings in Table~\ref{tab:ranking_metrics}. Fine-tuning on our dataset improves all metrics, and motion augmentation further improves results except for roughness appearance.

\paragraph{User Study.}
We conduct a user study to validate the accuracy of our VLM-based evaluation. We recruit 25 CG experts and use a pairwise preference test between DiffusionRenderer and our fine-tuned model, since judging material cues in complex scenes requires domain expertise. For metallic and roughness, we sample three cases where the VLM prefers DiffusionRenderer and three where it prefers ours, and report the agreement rate between experts and the VLM across questions. As shown in Table~\ref{tab:user_study}, expert judgments generally align with the VLM, with lower agreement for roughness due to more ambiguous cues.

\begin{figure*}[t] 
    \centering
    \includegraphics[width=\textwidth]{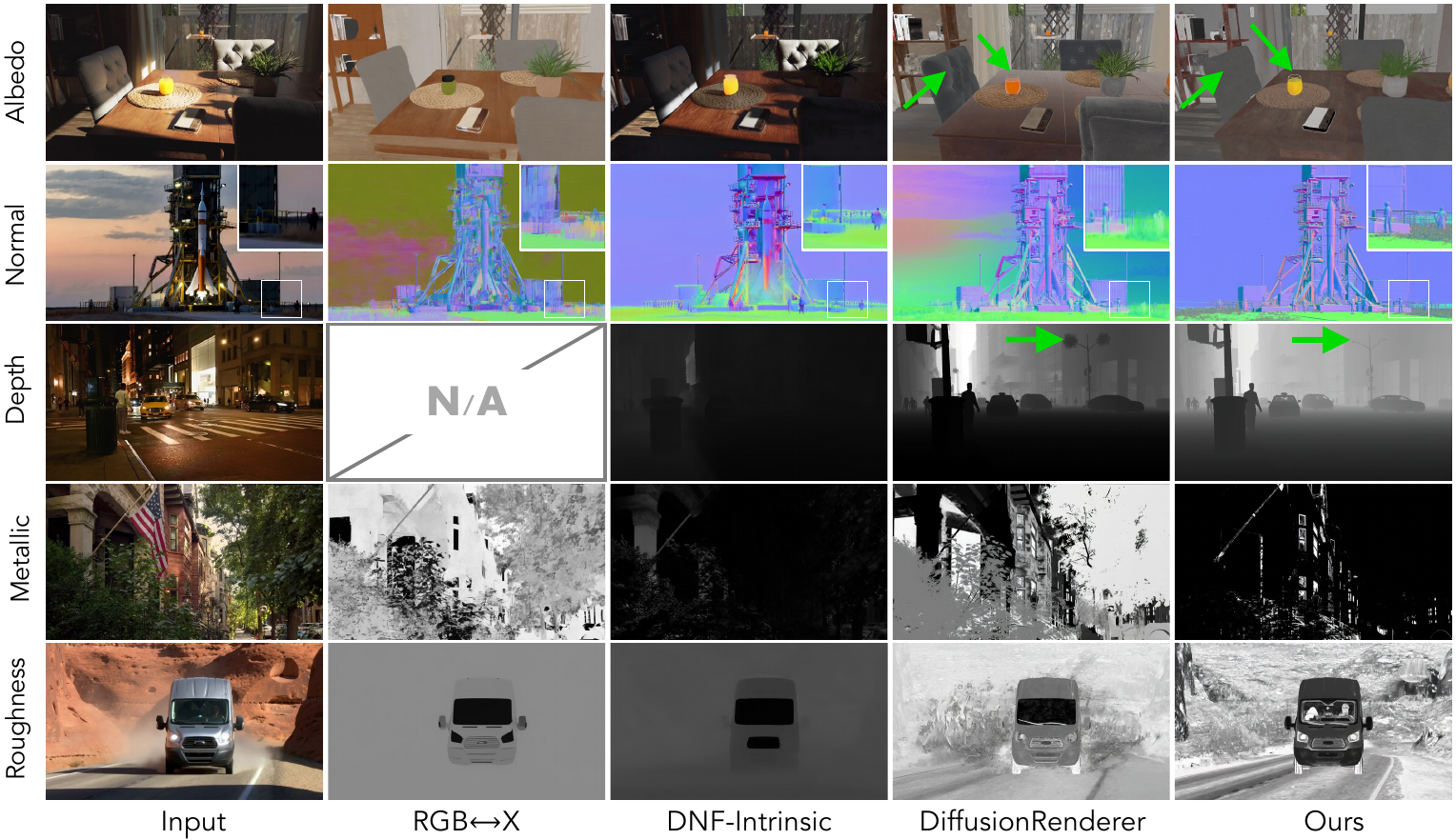}
    \caption{\textbf{Qualitative comparison of inverse rendering on in-the-wild data.} (Top to bottom: albedo, normal, depth, metallic, roughness). Our method significantly outperforms DiffusionRenderer in inverse rendering. It produces cleaner albedo with better delighting, artifact-free geometry, and robust material predictions that effectively resist complex outdoor illumination and atmospheric disruptions like smoke.}
    \label{fig:visual_results}
\end{figure*}

\subsection{Qualitative Evaluation of Inverse Rendering}
\begin{figure*}[t] 
    \centering
    \includegraphics[width=\textwidth]{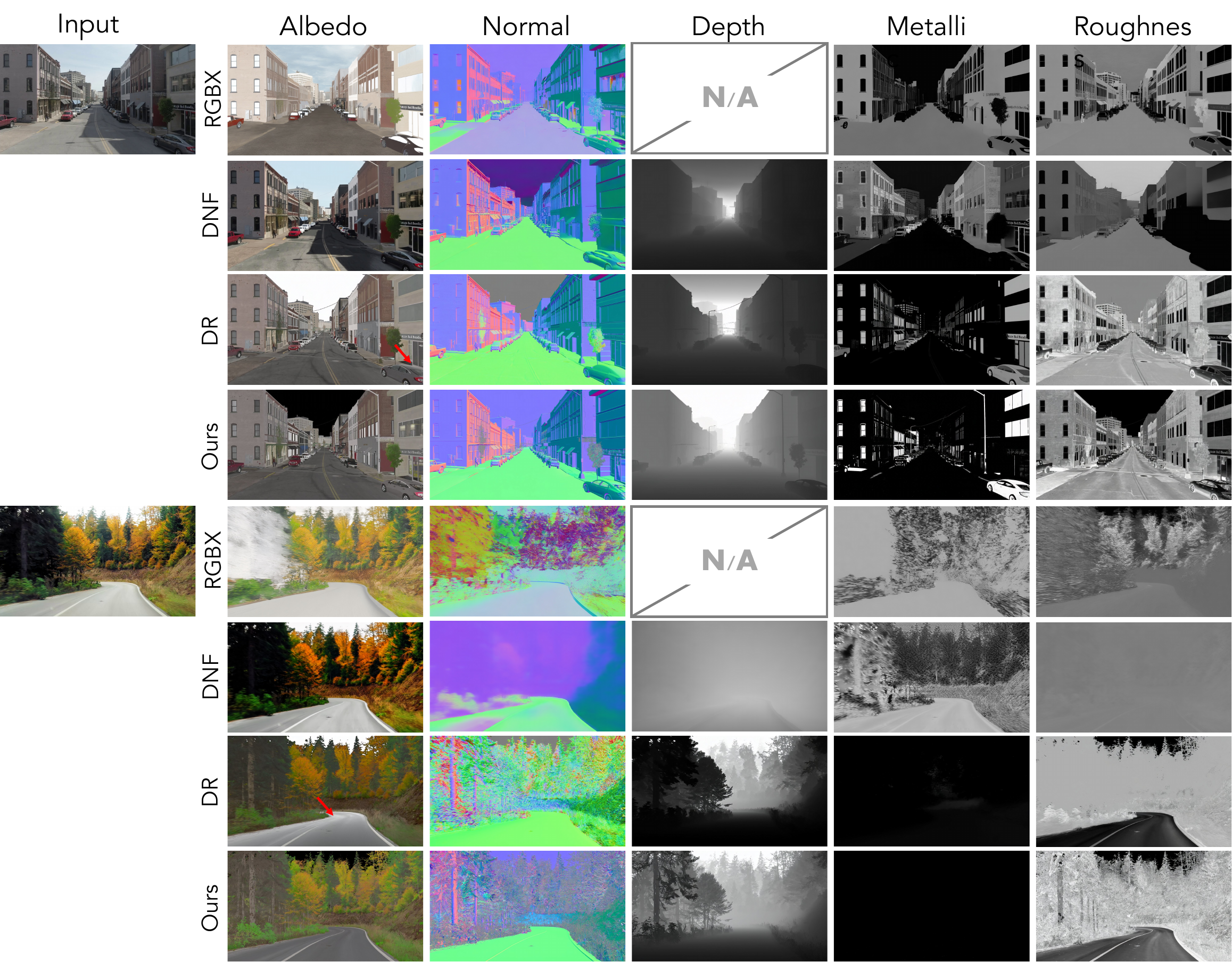}
    \caption{\textbf{Qualitative comparison of inverse rendering on in-the-wild data.}
    }
    \label{fig:visual_results_all_gbuffers}
\end{figure*}

We present qualitative comparisons on real-world video sequences in Figure~\ref{fig:visual_results}, visualizing (top to bottom) albedo, normals, depth, metallic, and roughness. Compared to DiffusionRenderer, our method exhibits superior performance in disentangling intrinsic scene properties. Notably, it produces highly clean albedo maps with thorough delighting, and reconstructs more precise depth and normals that faithfully preserve structures while resisting outdoor illumination artifacts. Furthermore, our fine-tuned model yields semantically accurate metallic and roughness predictions, successfully overcoming transient atmospheric disruptions like smoke and volumetric scattering. These visual results not only corroborate our quantitative findings but also explicitly highlight the critical advantage of our proposed dataset in enabling robust, physically grounded inverse rendering in the wild. More visualizations are detailed in Figure~\ref{fig:visual_results_all_gbuffers}.

\subsection{Ablation Study of Inverse Renderer}
\begin{figure}[!t]
    \centering
    \includegraphics[width=0.7\columnwidth]{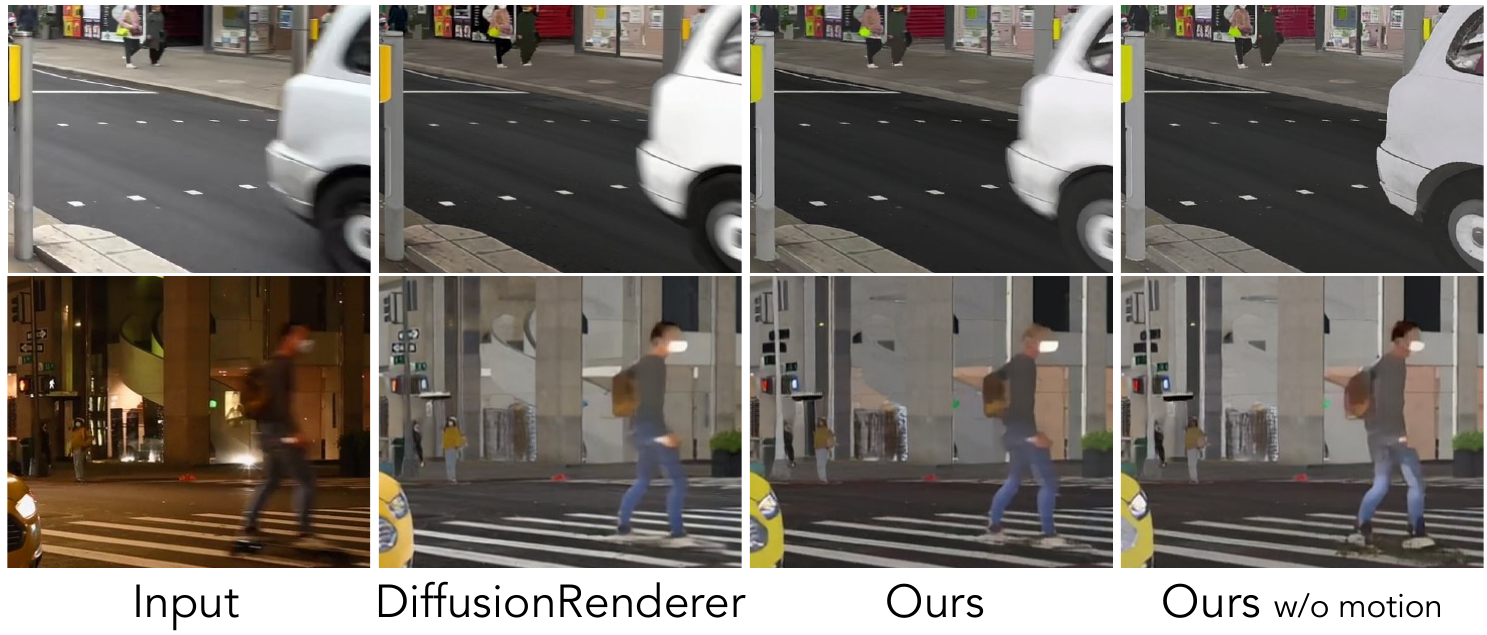}
    \caption{\textbf{Visual results of ablation study.}}
    \label{fig:ablation_study}
\end{figure}

We ablate motion effects by fine-tuning two variants under identical settings: with vs.\ without motion augmentation. Motion augmentation improves most synthetic metrics (Table~\ref{tab:ablation_mb}) and yields better real-video generalization with higher temporal stability (Figure~\ref{fig:ablation_study}), especially under strong motion blur where it reduces flicker and boundary crawling.

\subsection{Evaluation of Relighting}
\begin{figure*}[t] 
    \centering
    \includegraphics[width=\textwidth]{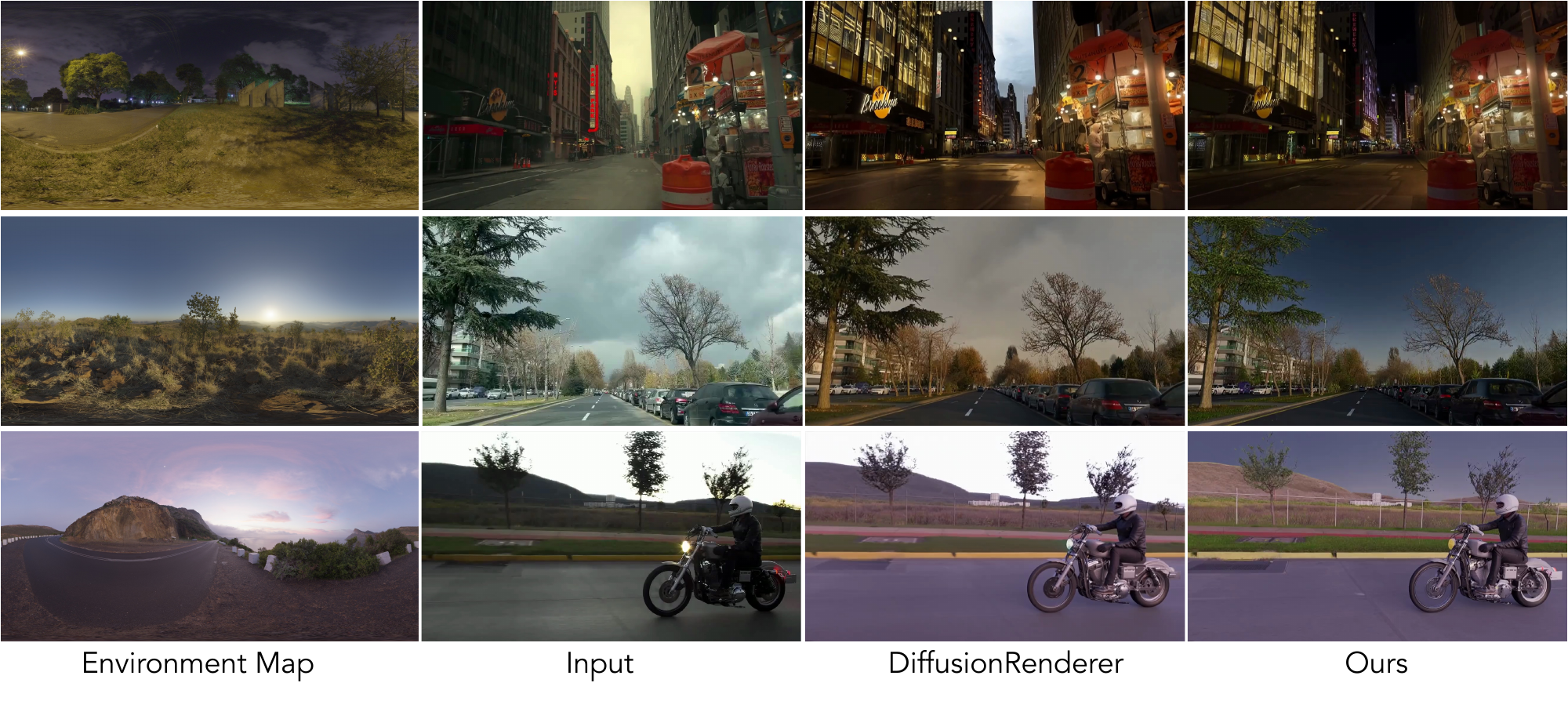}
    \caption{\textbf{Application of relighting on in-the-wild data.} 
    }
    \label{fig:relighting_results}
\end{figure*}

We further conduct a qualitative evaluation of relighting to demonstrate the generalization benefits unlocked by our dataset. Specifically, we collect diverse environment maps and synthesize images using the frozen forward renderer of DiffusionRenderer, conditioned on the G-buffers estimated by both the baseline and our fine-tuned inverse renderer. As shown in Figure~\ref{fig:relighting_results}, although the forward renderer is not fine-tuned on our dataset, the images synthesized from our improved G-buffers exhibit significantly better consistency with the target environment maps, particularly in the sky regions where baseline models often struggle. We attribute this superior relighting to our data-centric paradigm's enhanced ability to decouple intrinsic scene properties from environmental illumination. By disentangling lighting effects and spurious highlights during the inverse rendering stage, the fine-tuned model yields exceptionally clean and accurate G-buffers. Consequently, this enables the off-the-shelf forward renderer to generate highly realistic, illumination-consistent novel views, proving the efficacy of our proposed data. In essence, these results highlight that scaling and improving training data is a highly promising, direct pathway to overcoming the inherent ambiguities of inverse rendering in the wild.

\subsection{Evaluation on Game Editing}

\begin{figure*}[t] 
    \centering
    \includegraphics[width=\textwidth]{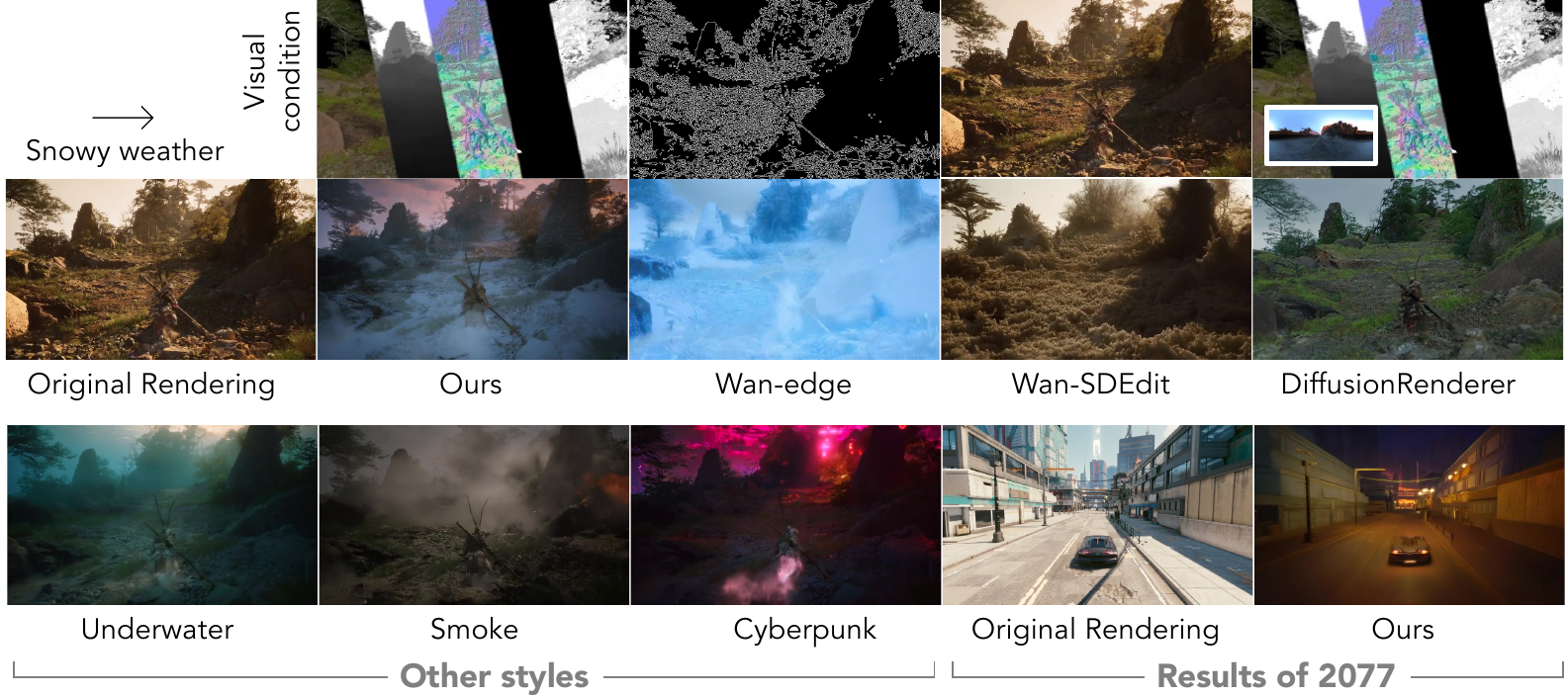}
    \caption{\textbf{Application of game editing.} We can craft the game styles (\eg, lighting, weather, and visual effect) using the G-buffers from an AAA game as conditions. 
    }
    \label{fig:game_results}
\end{figure*}

To showcase the practical efficacy of our dataset for downstream tasks, we investigate its application in high-fidelity video game editing. As illustrated by the qualitative comparisons in Figure~\ref{fig:game_results}, our G-buffer-conditioned model, fine-tuned on our data, outperforms existing video-to-video baselines. We benchmark our method against three representative paradigms: (i) a ControlNet-based framework guided by RGB-derived edge maps; (ii) an SDEdit-style stochastic editing pipeline; and (iii) a physics-informed baseline leveraging DiffusionRenderer, where the environment maps are estimated from our outputs via DiffusionLight. To ensure a rigorous and controlled comparison, all diffusion-based baselines utilize the same pre-trained backbone and text prompts as our method.

Compared to these alternatives, our approach achieves a superior balance between editability and visual fidelity to the original game render. Relying solely on edge maps provides spatial conditioning that preserves basic geometry but fails to guarantee material fidelity. Furthermore, because edge extraction from raw RGB frames is inherently unstable, this baseline suffers from severe temporal inconsistencies in the output video. In the context of video games, however, high-quality G-buffers can be stably retrieved as a native byproduct, serving as a superior alternative to noisy, image-derived proxies. Conversely, SDEdit introduces excessive deviation from the input; crucial but visually small objects frequently disappear, which disrupts underlying gameplay logic and degrades player interactivity. While DiffusionRenderer successfully maintains both geometry and material properties, it struggles with aggressive editing tasks, such as dramatic style transfers or the insertion of novel visual effects. Additionally, its reliance on environment maps severely limits user accessibility. In contrast, because our training data inherently captures complex in-game visual effects (e.g., volumetric fog, rain), our model learns a flexible prior. Rather than being strictly bottlenecked by the bare input geometry, it can seamlessly hallucinate and integrate rich atmospheric effects during the editing process.






\section{Conclusion}
In this work, we present a large-scale video dataset designed to unify inverse and forward rendering. Curated from high-fidelity commercial games, it provides long-term RGB sequences synchronized with dense G-buffers, alongside synthesized motion-blur variants to bridge the sim-to-real gap. To tackle the challenge of evaluating in-the-wild performance without ground truth, we introduce a novel, scalable VLM-based evaluation protocol. Extensive experiments show that fine-tuning DiffusionRenderer on our dataset substantially improves both robust material decomposition and the fidelity of G-buffer-conditioned video synthesis. Ultimately, our data and training recipe enable highly reliable, temporally coherent bidirectional rendering, offering a critical foundation for advanced world simulation and controllable generative editing in the wild.

\clearpage
\section*{Appendix}
\renewcommand{\thesection}{A}
\renewcommand{\thesubsection}{\thesection.\arabic{subsection}}
\renewcommand{\thefigure}{\thesection.\arabic{figure}}
\renewcommand{\thetable}{\thesection.\arabic{table}}

\setcounter{figure}{0}
\setcounter{table}{0}


\subsection{License Statement and Data Release Policy}
\label{sec:license}

Our dataset is constructed using in-game visual and geometric data from two commercial titles: \textit{Cyberpunk 2077} (CD PROJEKT RED) and \textit{Black Myth: Wukong} (Game Science). To ensure strict ethical and legal compliance with the developers' intellectual property rights, we outline our data collection methodology, release policy, and licensing terms as follows:

\begin{itemize}
    \item \textbf{Data Collection Methodology:} To ensure strict legal compliance with the End User License Agreements (EULA), our data curation toolkit operates entirely at the rendering API level. Our pipeline strictly intercepts the graphics API to capture runtime G-buffers (e.g., albedo, normals, depth) and final rendering outputs during gameplay. It does \textit{not} involve decompiling the game executables, circumventing anti-tamper mechanisms, or extracting proprietary source assets (such as original 3D meshes or textures) from the games' installation files. \textbf{This API-level capture aligns with established fair-use practices for dataset curation in the computer vision community.}
    \item \textbf{Compliance and Licensing:} In accordance with the developers' Fan Content Policies and EULAs, which permit non-commercial derivative works and sharing, our dataset will be released under the \textbf{Creative Commons Attribution-NonCommercial-ShareAlike 4.0 International (CC BY-NC-SA 4.0)} license. This strictly limits the usage of our dataset to non-commercial research purposes.
\item \textbf{Precedents:} Our data collection and distribution framework is highly consistent with established and widely recognized synthetic datasets in the computer vision community, such as the GTA-V dataset~\cite{richter2016playing} and VIPER~\cite{richter2017playing}, which have profoundly catalyzed subsequent breakthroughs in various downstream computer vision tasks.
    \item \textbf{Gated Access:} To prevent unauthorized mass distribution, we will not provide direct, public download links. Instead, the dataset will be released via gated access. Researchers wishing to use the dataset must formally agree to and sign a strict Terms of Use (ToU) agreement, acknowledging the original copyrights and committing to non-commercial use, before access is granted.
    \item \textbf{Open-Source Toolkit:} To promote transparency, reproducibility, and future research in video inverse rendering and relighting, we will fully open-source our data curation toolkit. This will enable researchers to utilize our pipeline to legally curate data from other games, facilitating the continuous expansion and diversification of such datasets.
\end{itemize}

\subsection{VLM Evaluation Prompt}
\label{sec:prompt}

\begin{figure*}[t]
    \centering
    \includegraphics[width=\textwidth]{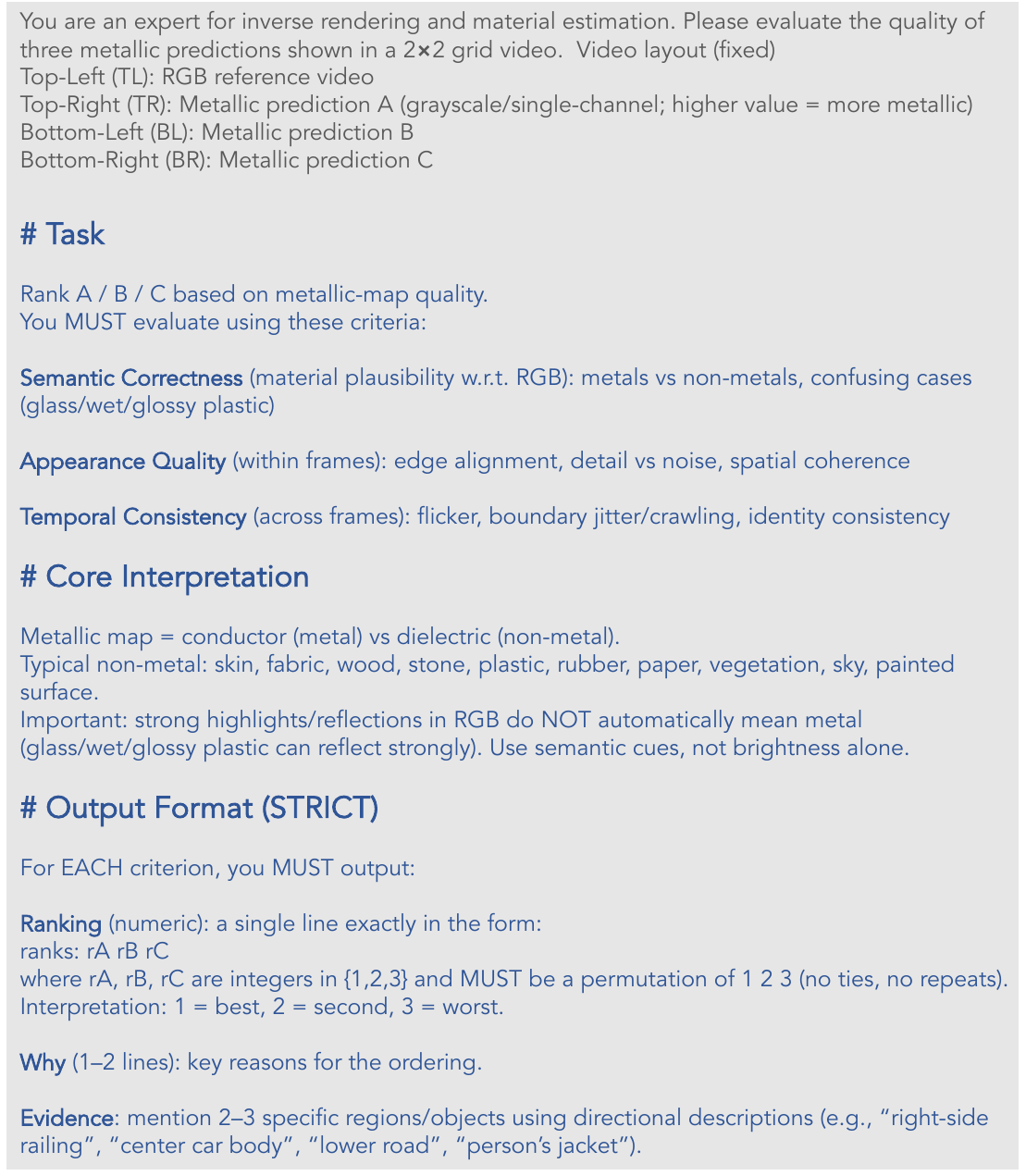}
    \caption{\textbf{Prompt used for VLM evaluation of metallic prediction.}}
    \label{fig:prompt}
\end{figure*}

To ensure reproducibility, we provide the exact prompt used for the Vision-Language Model (VLM) evaluation regarding metallic prediction. The details of the prompt design are illustrated in Figure~\ref{fig:prompt}.

\clearpage
\bibliographystyle{abbrv}
\bibliography{sample-base}

\clearpage

\end{document}